\documentclass[conference]{IEEEtran}
\IEEEoverridecommandlockouts

\usepackage{cite}
\usepackage{amsmath,amssymb,amsfonts}
\usepackage{algorithmic}
\usepackage{algorithm}
\usepackage{graphicx}

\usepackage{textcomp}
\usepackage{xcolor}
\usepackage{booktabs}
\usepackage{multirow}
\usepackage{url}
\usepackage[compatibility=false]{caption}
\usepackage{subcaption}
\usepackage{orcidlink}
\usepackage{dblfloatfix}
\usepackage{tikz}
\usepackage{pgfplots}

\makeatletter
\newcommand{\linebreakand}{%
  \end{@IEEEauthorhalign}
  \hfill\mbox{}\par
  \mbox{}\hfill\begin{@IEEEauthorhalign}
}
\makeatother

\title{SAM-Enhanced Segmentation on Road Datasets: Balancing Critical Classes in Autonomous Driving}

\author{\small Toomas Tahves$^{1}$, Mauro Bellone$^{2,3}$, Junyi Gu$^{4,5}$, Raivo Sell$^{1}$ \\ 
$^{1}$\textit{Department of Mechanical and Industrial Engineering, Tallinn University of Technology, Estonia} \\ 
$^{2}$\textit{FinEst Centre for Smart Cities, Tallinn University of Technology, Estonia} \\
$^{3}$\textit{Department of Computer Science and Engineering, Universitas Mercatorum, Rome, Italy} \\
$^{4}$\textit{Department of Computer Science and Engineering, Chalmers University of Technology, Gothenburg, Sweden} \\
$^{5}$\textit{University of Gothenburg, SE-412 96, Sweden } \\
}

\begin{document}

\maketitle

\begin{abstract}
Dense semantic segmentation is essential for autonomous driving, yet many multi-modal datasets lack pixel-level annotations. 
The Zenseact Open Dataset (ZOD) provides rich multi-sensor data but only bounding-box labels, limiting its use for segmentation research. 
Our primary contribution is a Segment Anything Model (SAM)-based annotation pipeline that produces dense, pixel-level annotations for ZOD by converting bounding boxes into semantic masks. 
In this pilot study, we process over 100{,}000 frames and manually curate a 2{,}300-frame subset (36\% acceptance rate) to establish a reliable baseline. 
Using these annotations, we evaluate transformer-based CLFT and CNN-based DeepLabV3+ architectures across diverse weather conditions, achieving up to 48.1\% mIoU with CLFT-Hybrid. 
To address extreme class imbalance, where pedestrians, cyclists, and signs constitute less than 1\% of pixels, we explore specialized models targeting rare classes. 
We further validate the pipeline on the Iseauto autonomous-vehicle platform, achieving 77.5\% mIoU, and show that SAM-derived representations transfer effectively across sensor configurations via bidirectional transfer learning. 
All code and annotations are released to support reproducible research.
\end{abstract}

\begin{IEEEkeywords}
autonomous driving, semantic segmentation, multi-modal fusion, Vision Transformers, DeepLabV3+, model specialization, class imbalance, transfer learning, computational efficiency, Segment Anything Model
\end{IEEEkeywords}

\section{Introduction}

Pixel-level semantic segmentation is a core component of autonomous-vehicle perception, enabling fine-grained scene understanding in complex environments \cite{janai2020computer}. 
Early convolutional architectures such as U-Net\cite{unet}, PSPNet\cite{pspnet}, and DeepLabV3+\cite{deeplabv3plus} established strong baselines for dense prediction, while recent Vision Transformer models\cite{vit,liu2021swin,cheng2021mask2former,segformer,setr} offer improved global context modeling at the cost of higher computational demand. 
Despite these advances, multi-modal segmentation still requires large-scale datasets with dense, high-quality annotations across diverse environmental conditions.

The Zenseact Open Dataset (ZOD)\cite{zod} provides rich camera–LiDAR data from Northern European environments, complementing established datasets such as nuScenes\cite{caesar2020nuscenes} and KITTI\cite{geiger2012we}. 
However, ZOD provides only 2D bounding boxes for traffic participants, leaving a gap for supervised segmentation research that requires pixel-level labels.

To reduce annotation cost, automated and semi-automated labeling approaches have been explored. 
The Segment Anything Model (SAM)\cite{sam} enables large-scale mask generation without task-specific training, yet its outputs may contain hallucinations, boundary errors, or failures on small or occluded objects, especially in safety-critical domains. 
This exposes an annotation-method gap: foundation-model outputs require filtering, validation, and structured integration to produce reliable dense labels.

Multi-modal perception has been studied through early, late, and feature-level fusion strategies, with feature-level fusion often yielding superior performance\cite{hazirbas2016fusenet}, particularly in adverse weather where LiDAR remains reliable. 
However, autonomous-driving datasets exhibit extreme class imbalance, and standard loss-balancing techniques such as class-balanced loss\cite{cui2019class} and focal loss\cite{focal_loss} often struggle to emphasize rare but safety-critical classes. 
Model specialization has shown promise in object detection\cite{cascade_rcnn}, but its potential for dense multi-modal segmentation, especially when combined with SAM-based annotations, remains largely unexplored, revealing a model-design gap.

To address these dataset, annotation, and model gaps, we develop a SAM-based preprocessing pipeline that converts ZOD bounding boxes into dense masks. 
In this pilot study, we process over 100{,}000 frames and manually curate a 2{,}300-frame subset to establish a reliable benchmark. 
Using these annotations, we evaluate segmentation performance across diverse weather conditions using CLFT models\cite{clft1,clfcn} and DeepLabV3+, investigate rare-class challenges, and assess cross-dataset generalization on the Iseauto autonomous-vehicle platform through bidirectional transfer learning.

Our contributions are as follows:
\begin{itemize}
    \item A SAM-based annotation pipeline that produces dense masks for multi-modal semantic segmentation.
    \item A curated pilot dataset of 2{,}300 manually verified frames enabling the first segmentation benchmark on ZOD.
    \item A comparative evaluation of CLFT and DeepLabV3+ architectures under diverse weather conditions.
    \item Class-specialized models that improve segmentation of rare but safety-critical categories.
    \item Cross-dataset validation on the Iseauto platform and bidirectional transfer-learning analysis.
\end{itemize}

\begin{figure*}[ht]
\centering
\begin{minipage}[b]{0.32\textwidth}
\centering
\includegraphics[width=\textwidth]{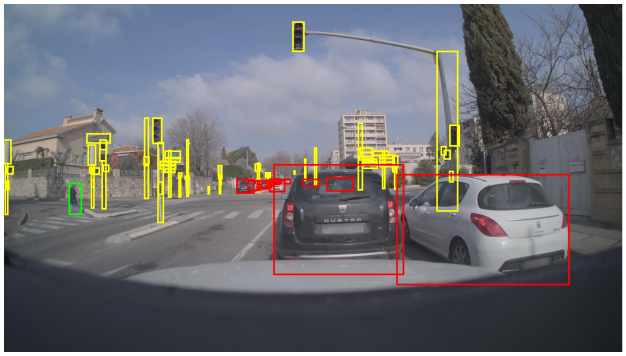}
\subcaption{Original ZOD bounding boxes showing overlapping and duplicate boxes}
\label{fig:sam_pipeline_a}
\end{minipage}
\hfill
\begin{minipage}[b]{0.32\textwidth}
\centering
\includegraphics[width=\textwidth]{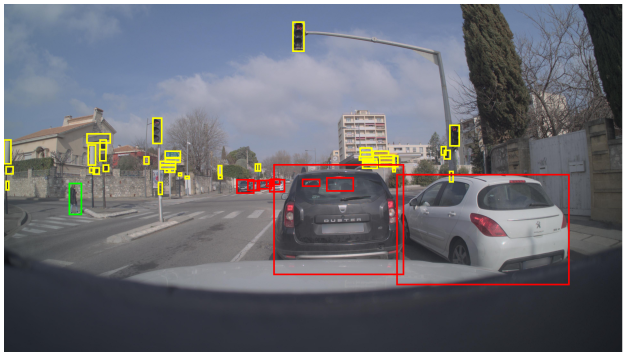}
\subcaption{Clean bounding boxes after filtering and deduplication}
\label{fig:sam_pipeline_b}
\end{minipage}
\hfill
\begin{minipage}[b]{0.32\textwidth}
\centering
\includegraphics[width=\textwidth]{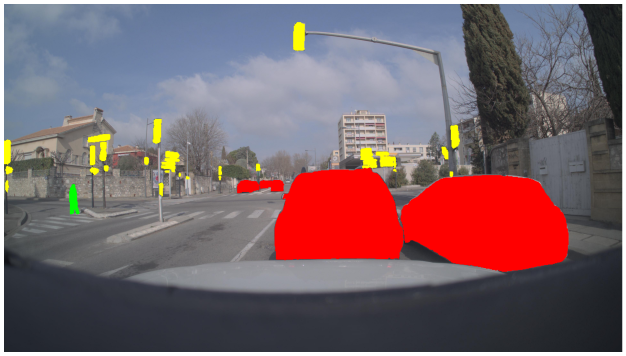}
\subcaption{SAM semantic segmentation masks generated from filtered bounding boxes}
\label{fig:sam_pipeline_c}
\end{minipage}
\caption{SAM-based preprocessing pipeline visualization showing the progression from raw ZOD bounding boxes to dense semantic segmentation masks. (a) Original bounding boxes from ZOD dataset, (b) After filtering, deduplication, and size constraints, (c) Final pixel-level segmentation masks generated by SAM with priority-based resolution.}
\label{fig:sam_pipeline}
\end{figure*}

\begin{table*}[htbp]
\centering
\caption{ZOD and Iseauto Dataset Pixel Statistics by Weather Condition (384×384 Resolution)}
\label{tab:dataset_stats}
\begin{tabular}{@{}l l r r r r r @{}}
\toprule
Dataset & Weather & Sample & Background & Vehicle & Human & Sign \\
 & Condition & Size & \multicolumn{4}{c}{\scriptsize (pixels / \% of total)} \\
\midrule
\multirow{6}{*}{ZOD} & Day Fair & 1,421 & 204.4M / 97.6\% & 3.4M / 1.6\% & 733.4K / 0.4\% & 912.7K / 0.4\% \\
 & Day Rain & 234 & 33.6M / 97.4\% & 643.5K / 1.9\% & 86.2K / 0.3\% & 167.1K / 0.5\% \\
 & Night Fair & 385 & 55.9M / 98.5\% & 558.7K / 1.0\% & 46.2K / 0.1\% & 231.9K / 0.4\% \\
 & Night Rain & 99 & 14.4M / 98.6\% & 147.5K / 1.0\% & 6.7K / 0.1\% & 52.0K / 0.4\% \\
 & Snow & 161 & 23.3M / 98.2\% & 276.3K / 1.2\% & 18.7K / 0.1\% & 89.3K / 0.4\% \\
 & \textbf{Total} & \textbf{2,300} & \textbf{331.7M / 97.8\%} & \textbf{5.1M / 1.5\%} & \textbf{891.4K / 0.3\%} & \textbf{1.5M / 0.4\%} \\
\midrule
\multirow{5}{*}{Iseauto} & Day Fair & 600 & 85.6M / 96.7\% & 2.3M / 2.6\% & 606.8K / 0.7\% & 0 / 0\% \\
 & Day Rain & 600 & 84.5M / 95.5\% & 3.8M / 4.3\% & 176.0K / 0.2\% & 0 / 0\% \\
 & Night Fair & 600 & 85.4M / 96.6\% & 2.7M / 3.1\% & 290.5K / 0.3\% & 0 / 0\% \\
 & Night Rain & 600 & 86.7M / 98.0\% & 1.7M / 1.9\% & 121.4K / 0.1\% & 0 / 0\% \\
 & \textbf{Total} & \textbf{2,400} & \textbf{342.2M / 96.7\%} & \textbf{10.5M / 3.0\%} & \textbf{1.2M / 0.3\%} & \textbf{0 / 0\%} \\
\bottomrule
\end{tabular}
\end{table*}

\begin{figure}[ht]
\centering
\includegraphics[width=\columnwidth]{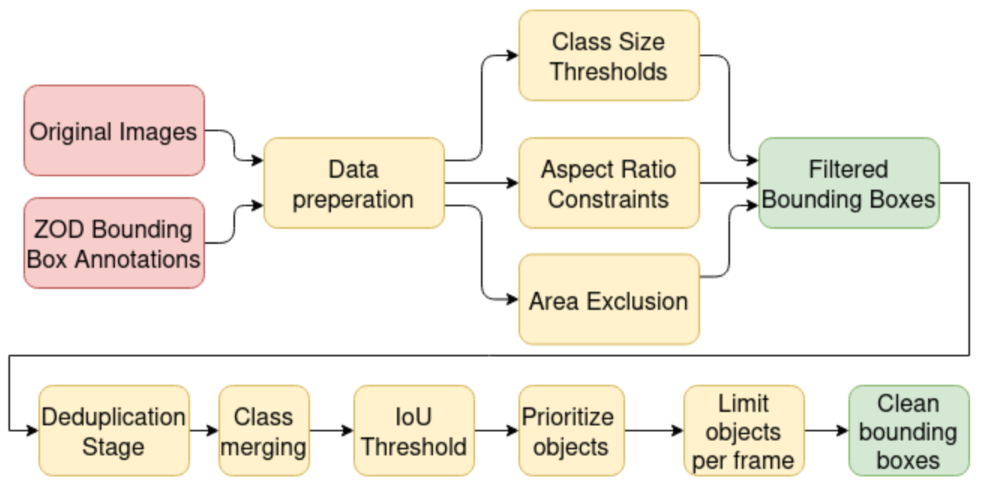}
\caption{Data preprocessing pipeline for converting ZOD bounding boxes to dense segmentation masks. The process includes input filtering with class-specific size thresholds, aspect ratio constraints, and area exclusion; deduplication through intra-class merging with IoU threshold of 0.3 while prioritizing larger objects; and limiting to 75 objects per frame to optimize processing efficiency and annotation quality.}
\label{fig:data_preprocess}
\end{figure}

\section{Methodology}
\subsection{SAM-based Preprocessing for ZOD}
The Zenseact Open Dataset (ZOD) provides 2D bounding-box annotations for object detection but lacks dense pixel-level labels for semantic segmentation. 
To enable supervised training, we convert these bounding boxes into dense masks using the Segment Anything Model (SAM). 
From more than 100{,}000 processed frames, we manually inspected 6{,}400 SAM outputs and selected 2{,}300 masks (36\% acceptance rate) that were visually consistent and free of major artifacts. 
SAM is known to struggle with small, distant, or heavily occluded objects, and often produces boundary drift or spurious regions in cluttered scenes \cite{sam,kirillov2024sam2}.
Table~\ref{tab:dataset_stats} summarizes the resulting dataset statistics, and Figure~\ref{fig:sam_pipeline} illustrates the transformation from noisy bounding boxes to refined dense masks.

Our processing pipeline consists of the following stages, summarized in Algorithm~\ref{alg:sam_pipeline}:

\begin{algorithm}
\small
\caption{SAM-based Preprocessing Pipeline}
\label{alg:sam_pipeline}
\begin{algorithmic}[1]
\STATE \textbf{Input:} Bounding boxes $B$, image $I$ of size $H \times W$
\STATE \textbf{Output:} Segmentation mask $M \in \{0,1,\dots,C\}^{768 \times 768}$
\vspace{0.15cm}
\STATE $B_{\text{filtered}} \leftarrow \text{BoundingBoxFiltering}(B, H, W)$   \hspace{0.2cm} Algorithm~\ref{alg:sam_stage1}

\STATE $M_{\text{sam}} \leftarrow \text{SAMInference}(I, B_{\text{filtered}})$ \hspace{1.5cm}Algorithm~\ref{alg:sam_stage2}

\STATE Resize $M_{\text{sam}}$ to $M \in \{0,1,\dots,C\}^{768 \times 768}$ 

\STATE \textbf{return} $M$
\end{algorithmic}
\end{algorithm}

1. Bounding‑Box Filtering and Deduplication: We begin by filtering the raw ZOD bounding boxes to remove detector noise and ensure that only plausible object candidates are passed to SAM (Algorithm \ref{alg:sam_stage1}).
Class‑specific minimum size thresholds (vehicles $>30$ px signs, pedestrians, cyclists $>15$ px) remove extremely small detections unlikely to yield reliable masks.
Aspect‑ratio constraints ($<8:1$) eliminate elongated artifacts, and boxes covering more than 40\% of the image are excluded to avoid dominant detections that could bias SAM outputs.
Thresholds were selected to balance annotation quality and dataset coverage based on preliminary visual inspection.

To reduce redundancy, we merge overlapping boxes within each class using IoU‑based deduplication (IoU $<$ 0.3), retaining the larger box when conflicts occur.
Finally, we cap each frame at the 75 largest boxes to balance completeness with computational efficiency. This stage is visualized in Figure
\ref{fig:data_preprocess}.

\begin{algorithm}
\small
\caption{Bounding Box Filtering and Deduplication}
\label{alg:sam_stage1}
\begin{algorithmic}[1]
\STATE \textbf{Input:} Bounding boxes $B = \{b_i\}_{i=1}^n$, image size $H \times W$
\STATE \textbf{Output:} Filtered bounding boxes $\mathcal{B}_{\text{filtered}}$
\vspace{0.15cm}
\STATE $\mathcal{B}_{\text{valid}} \leftarrow \emptyset$

\FOR{each bounding box $b \in B$}
    \STATE Extract width $w_b$, height $h_b$, class $c_b$
    \IF{validity check} 
        \STATE Add $b$ to $\mathcal{B}_{\text{valid}}$
    \ENDIF
\ENDFOR

\STATE $\mathcal{B}_{\text{merged}} \leftarrow \text{IoU\_deduplicate}(\mathcal{B}_{\text{valid}}, \tau = 0.3)$

\STATE $\mathcal{B}_{\text{filtered}} \leftarrow \text{select\_top\_k}(\mathcal{B}_{\text{merged}}, k = 75)$

\STATE \textbf{return} $\mathcal{B}_{\text{filtered}}$
\end{algorithmic}
\end{algorithm}

2. SAM Inference with Priority‑Based Mask Fusion: SAM inference is performed on a resized 1024 px image for efficiency (Algorithm \ref{alg:sam_stage2}).
Each filtered bounding box is used as a prompt for SAM’s ViT‑H model, processed in batches of 16 for throughput.
SAM produces a binary mask for each object.

Because SAM masks may overlap, we resolve conflicts using a class‑priority scheme: pedestrians and cyclists receive the highest priority, followed by signs and vehicles.
When two masks assign different labels to the same pixel, the label with higher priority overwrites the other.
This ensures that vulnerable road users are not occluded by larger objects in ambiguous regions.
The result is an intermediate semantic mask $M_{\text{sam}}$ at 1024 px resolution.

\begin{algorithm}
\small
\caption{SAM Inference with Priority-Based Mask Fusion}
\label{alg:sam_stage2}
\begin{algorithmic}[1]
\STATE \textbf{Input:} Image $I$, filtered bounding boxes $\mathcal{B}_{\text{filtered}}$
\STATE \textbf{Output:} Intermediate semantic mask $M_{\text{sam}}$
\vspace{0.15cm}
\STATE Resize image: $I' \leftarrow \text{resize}(I, \leq 1024 \times 1024)$
\STATE Initialize semantic mask $M_{\text{sam}} \in \{0,1,\dots,C\}^{1024 \times 1024}$ with background label $0$

\STATE Define class priority function $p: \mathcal{C} \rightarrow \mathbb{N}$

\FOR{each batch $\mathcal{B}_{\text{batch}} \subseteq \mathcal{B}_{\text{filtered}}$ of size 16}
    \STATE $\mathcal{M}_{\text{batch}} \leftarrow \emptyset$
    \FOR{each bounding box $b \in \mathcal{B}_{\text{batch}}$}
        \STATE $m_b \leftarrow \text{SAM.predict}(I', b, \text{multimask\_output}=\text{False})$
        \STATE Store $(b, m_b)$ in $\mathcal{M}_{\text{batch}}$
    \ENDFOR

    \FOR{each $(b, m_b) \in \mathcal{M}_{\text{batch}}$}
        \STATE Let $c_b$ be the class label of $b$, and $p_b \leftarrow p(c_b)$
        \FOR{each pixel $(u,v)$ such that $m_b(u,v) > 0$}
            \IF{$M_{\text{sam}}(u,v) = 0 \;\lor\; p_b \geq p(M_{\text{sam}}(u,v))$}
                \STATE $M_{\text{sam}}(u,v) \leftarrow c_b$
            \ENDIF
        \ENDFOR
    \ENDFOR
\ENDFOR

\STATE \textbf{return} $M_{\text{sam}}$
\end{algorithmic}
\end{algorithm}

\begin{figure*}[htbp]
\centering
\begin{minipage}[b]{0.32\textwidth}
\centering
\includegraphics[width=\textwidth]{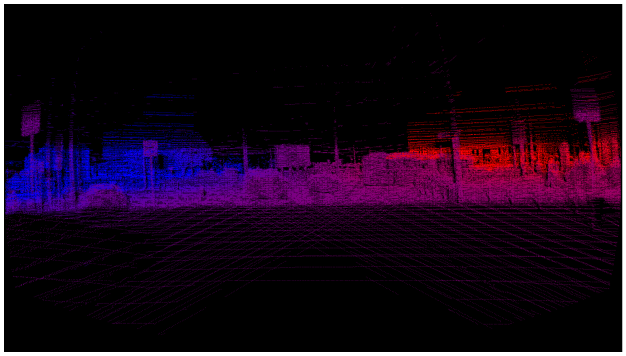}
\subcaption{LiDAR point cloud 2D projection with distance-based coloring (blue=near, red=far)}
\label{fig:lidar_pipeline_a}
\end{minipage}
\hfill
\begin{minipage}[b]{0.32\textwidth}
\centering
\includegraphics[width=\textwidth]{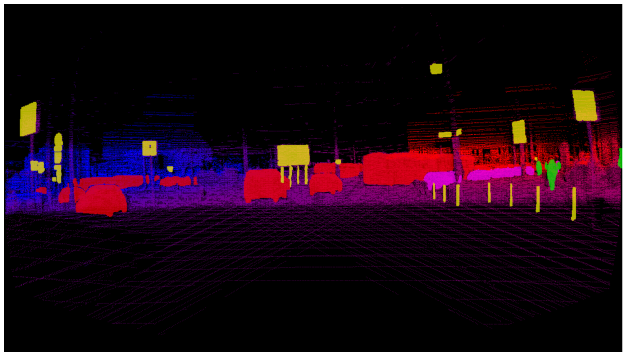}
\subcaption{SAM segmentation masks overlaid on LiDAR point cloud projection}
\label{fig:lidar_pipeline_b}
\end{minipage}
\hfill
\begin{minipage}[b]{0.32\textwidth}
\centering
\includegraphics[width=\textwidth]{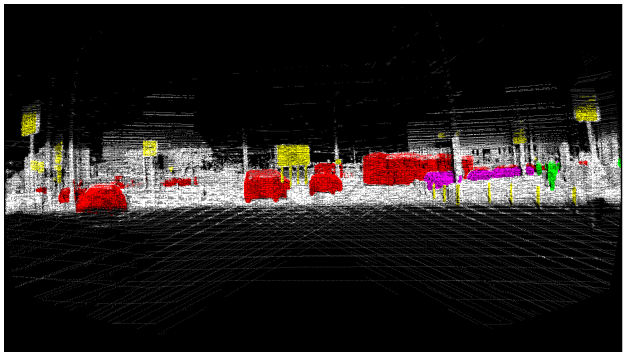}
\subcaption{LiDAR-native class annotations mapped to point cloud regions}
\label{fig:lidar_pipeline_c}
\end{minipage}
\caption{LiDAR annotation generation pipeline illustrating the creation of LiDAR-native segmentation masks using SAM guidance. (a) Raw LiDAR point cloud projection with depth-based coloring, (b) SAM segmentation masks overlaid on LiDAR data to identify object locations, (c) Final LiDAR-native annotations with class-specific coloring applied only to regions with LiDAR coverage.}
\label{fig:lidar_pipeline}
\end{figure*}

3. Output Generation: The fused SAM mask is resized to 768 px using bilinear interpolation (Algorithm \ref{alg:sam_pipeline}, line 5), producing the final dense segmentation mask used for training.
This converts ZOD’s sparse bounding boxes into pixel‑level annotations suitable for multi‑modal semantic segmentation.

For modality‑specific training, we derive two annotation variants from the SAM‑generated masks.
(1) Camera‑only annotations use the dense SAM masks with minimal filtering, preserving nearly the full camera field of view.
Only narrow border regions and very small components are removed to eliminate uncertain or noisy pixels.
(2) LiDAR‑only annotations project the SAM labels onto regions covered by actual LiDAR returns, applying distance filtering and strict geometric alignment to ensure that only points with valid sensor measurements receive labels.

\subsection{Camera-Only Annotation Generation}
To generate annotations suitable for camera‑only semantic segmentation, we adapt the SAM‑derived dense masks while introducing minimal ignore regions to account for areas of genuine uncertainty.
Unlike LiDAR‑based annotations, camera images provide full‑frame coverage without geometric sparsity, allowing us to retain nearly the entire field of view for training.

Starting from the dense SAM masks, we apply a conservative filtering strategy.
Only a narrow band along the image borders, approximately 1\% of the image height, is marked as an ignore region to account for potential cropping artifacts and peripheral distortions.
This ensures that the vast majority of the image remains available for supervision.
Additionally, very small connected components (fewer than 25 pixels), which are typically attributable to segmentation noise, are removed to improve label reliability.

The resulting annotations preserve full‑resolution camera information while excluding only those regions where label quality cannot be guaranteed.
This enables camera‑only models to learn from the complete visual scene without incurring loss penalties on ambiguous or low‑confidence areas.

\subsection{LiDAR-Only Annotation Generation}
To enable LiDAR‑only training and isolate modality‑specific performance, we generate LiDAR‑aligned annotations using real LiDAR returns combined with SAM‑derived object masks.
This provides supervision that reflects the inherent sparsity and geometric structure of LiDAR while avoiding the dense, camera‑driven masks used in the fusion setting.
As illustrated in Figure \ref{fig:lidar_pipeline}, the process projects SAM‑identified object regions onto the corresponding LiDAR points, ensuring that only points with actual sensor returns receive labels.

The generation pipeline consists of four steps:
(1) LiDAR coverage analysis to identify pixels containing valid LiDAR returns;
(2) distance filtering (up to 90m) to remove extremely sparse or noisy long‑range points;
(3) class assignment by intersecting SAM masks with LiDAR‑covered regions for vehicles, signs, pedestrians, and cyclists;
(4) quality checks to ensure strict geometric alignment without dilation or mask expansion.

This produces LiDAR‑native annotations that preserve the true sampling pattern of the sensor and allow training of LiDAR‑only segmentation models under conditions that reflect real‑world LiDAR sparsity.

\subsection{Iseauto Dataset Enhancement}
To improve annotation quality on the Iseauto dataset, originally labeled manually in prior work~\cite{clfcn}, we apply a SAM‑based edge refinement procedure.
This approach leverages SAM’s zero‑shot segmentation capability to correct boundary inaccuracies while preserving the semantic intent of the original labels.

Algorithm~\ref{alg:sam_edge_refinement} outlines the refinement process.
For each annotated object, SAM is prompted using the bounding‑box center to generate a high‑resolution proposal mask.
The refinement then proceeds conservatively along the object boundary.
(1) Edge Expansion adds pixels where SAM predicts object regions adjacent to the existing mask, helping fill small gaps or under‑segmented areas.
(2) Edge Contraction removes boundary pixels where SAM predicts background, reducing over‑segmentation.

Both operations are constrained by a maximum‑change threshold to prevent semantic drift and ensure that updates remain local.
This strategy improves geometric accuracy while maintaining consistency with the original human‑annotated class assignments.

\begin{algorithm}
\small
\caption{SAM Edge Refinement Algorithm}
\label{alg:sam_edge_refinement}
\begin{algorithmic}[1]
\STATE \textbf{Input:} Initial mask $M$, Image $I$, Bounding boxes $B$
\STATE \textbf{Output:} Refined mask $M_{refined}$
\vspace{0.15cm}
\STATE $M_{refined} \leftarrow M$
\FOR{each bounding box $b \in B$}
    \STATE $m_{sam} \leftarrow \text{SAM}(I, \text{center}(b))$
    
    \STATE \textbf{Edge Expansion:}
    \STATE Find background pixels adjacent to mask where $m_{sam}$ is object:
    \STATE $P_{add} \leftarrow (M_{refined}[b] == 0) \cap \text{dilate}(M_{refined}[b]) \cap (m_{sam} == 1)$
    \STATE Add $P_{add}$ to $M_{refined}[b]$ (limited by max\_changes)
    
    \STATE \textbf{Edge Contraction:}
    \STATE Find object pixels adjacent to background where $m_{sam}$ is background:
    \STATE $P_{rem} \leftarrow (M_{refined}[b] == c_b) \cap \text{dilate}(M_{refined}[b] == 0) \cap (m_{sam} == 0)$
    \STATE Remove $P_{rem}$ from $M_{refined}[b]$ (limited by max\_changes)
\ENDFOR
\STATE \textbf{return} $M_{refined}$
\end{algorithmic}
\end{algorithm}

\subsection{Network Architectures}
We evaluate two segmentation paradigms: Transformer‑based Camera–LiDAR Fusion Transformers (CLFT)~\cite{clft1,clfcn} and the CNN‑based DeepLabV3+~\cite{deeplabv3plus}.
CLFT (Base, Large, and Hybrid variants) employs a three‑stage encoder–decoder design with modality‑specific RGB and LiDAR encoders fused through cross‑modal multi‑head self‑attention. This structure enables joint reasoning over image appearance and LiDAR geometry at multiple feature scales.

DeepLabV3+ uses a ResNet‑101 backbone with atrous spatial pyramid pooling (ASPP) for multi‑scale context aggregation. For multi‑modal fusion, we extend the decoder to accept projected LiDAR features via channel concatenation, allowing the network to incorporate depth‑aligned geometric cues alongside RGB features.

\subsection{Evaluation Metrics}
Segmentation performance is measured using Intersection over Union (IoU), defined as the ratio between the intersection and union of predicted and ground‑truth masks for each class.
We report mean IoU (mIoU), computed as the unweighted average across classes, and Frequency‑Weighted IoU (FW IoU), which weights each class by its pixel frequency to reflect its prevalence in the dataset.

\section{Experiments and Results}

\subsection{Training Setup}
For our best-performing CLFT-Hybrid model, we used a batch size of 8 and a learning rate of $2 \times 10^{-4}$, decayed by 0.1 every 50 epochs.
Reported results are from single runs.
All models were initialized from ImageNet-pretrained weights to accelerate convergence.
The dataset was partitioned into a custom split of 50\% train, 25\% validation, and 25\% test to accommodate the curated subset of 2300 frames.
Splits were created by matching class‑pixel and weather distributions (rather than by raw frame counts), ensuring train/validation/test subsets share the same distribution.
Models were trained using the Adam optimizer with cross-entropy loss weighted to address class imbalance.
Data augmentation included random cropping, horizontal flipping, and color jittering. Models were trained on NVIDIA A100 GPUs (80GB) provided by the TalTech High Performance Computing Centre~\cite{herrmann}.

\subsection{Baseline Analysis on ZOD}
\begin{table}[htbp]
\centering
\caption{ZOD Dataset Baseline Performance Across Architectures, and Weather Conditions (best checkpoint IoU \%)}
\begin{tabular}{@{}l l c c c c c@{}}
\toprule
Architecture & Weather & mIoU & FW IoU & Vehicle & Sign & Human \\
\midrule
CLFT-Base & Day Fair & 44.5 & 55.8 & 66.0 & 32.3 & 35.2 \\
Fusion & Day Rain & 43.2 & 61.1 & 65.8 & 30.8 & 32.9 \\
 & Night Fair & 39.7 & 50.5 & 58.1 & 34.6 & 26.5 \\
 & Night Rain & 36.5 & 41.8 & 46.7 & 26.5 & 36.2 \\
 & Snow & 42.2 & 52.3 & 59.0 & 37.8 & 29.9 \\
\midrule
CLFT-Hybrid & Day Fair & \textbf{48.1} & \textbf{59.3} & \textbf{70.0} & \textbf{35.1} & \textbf{39.1} \\
Fusion & Day Rain & 42.9 & \textbf{62.7} & 65.4 & 29.7 & 33.6 \\
 & Night Fair & \textbf{47.5} & \textbf{56.5} & \textbf{66.0} & \textbf{36.8} & \textbf{39.8} \\
 & Night Rain & \textbf{42.7} & \textbf{51.8} & \textbf{58.9} & \textbf{33.8} & 35.3 \\
 & Snow & \textbf{47.2} & \textbf{59.3} & \textbf{67.4} & 39.1 & \textbf{35.0} \\
\midrule
CLFT-Large & Day Fair & 45.3 & 57.0 & 67.5 & 34.0 & 34.4 \\
Fusion & Day Rain & \textbf{45.3} & 62.2 & \textbf{67.1} & \textbf{34.5} & \textbf{34.2} \\
 & Night Fair & 41.1 & 51.9 & 59.7 & 35.3 & 28.4 \\
 & Night Rain & 36.1 & 39.9 & 43.9 & 27.2 & \textbf{37.3} \\
 & Snow & 45.4 & 57.1 & 64.7 & \textbf{40.7} & 30.8 \\
\midrule
DeepLabV3+ & Day Fair & 33.2 & 46.4 & 56.4 & 19.5 & 23.5 \\
Fusion & Day Rain & 37.4 & 50.5 & 60.3 & 22.1 & 29.7 \\
& Night Fair & 29.9 & 37.8 & 44.8 & 21.9 & 23.0 \\
& Night Rain & 28.2 & 36.5 & 42.6 & 18.4 & 23.6 \\
& Snow & 32.6 & 45.7 & 54.2 & 24.4 & 19.1 \\
\bottomrule
\end{tabular}
\label{tab:zod_baseline}
\end{table}

Table \ref{tab:zod_baseline} summarizes ZOD fusion performance under different weather conditions.
Across all settings, CLFT‑Hybrid achieves the strongest overall results, reaching 48.1\% mIoU in Day Fair and maintaining comparatively stable performance in adverse conditions.
In contrast, DeepLabV3+ exhibits larger performance drops, particularly at night.
The Transformer‑based CLFT variants show higher stability across weather.
One plausible explanation is that self‑attention can better integrate long‑range context and suppress isolated noisy returns, whereas CNNs may be more sensitive to local perturbations in LiDAR or image features.
However, performance on safety‑critical classes such as signs (29.7-39.1\% IoU) and humans (33.6-39.8\% IoU) remains challenging, likely due to the high density of small, distant objects in ZOD and the limited spatial resolution available for these categories.

\begin{figure*}[htbp]
\centering
\begin{minipage}[b]{0.32\textwidth}
\centering
\includegraphics[width=\textwidth]{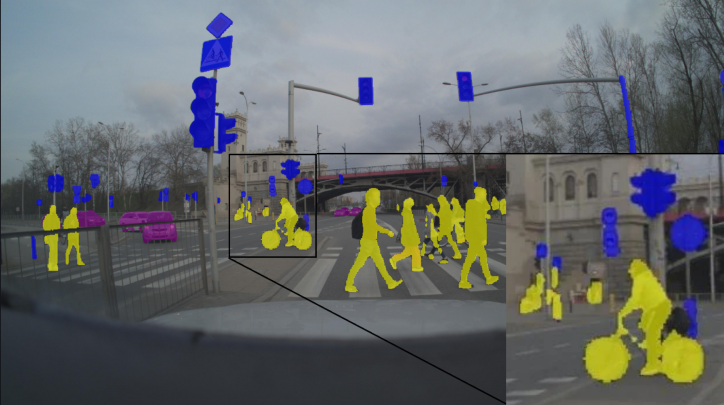}
\subcaption{SAM ground truth}
\label{fig:qualitative_gt}
\end{minipage}
\hfill
\begin{minipage}[b]{0.32\textwidth}
\centering
\includegraphics[width=\textwidth]{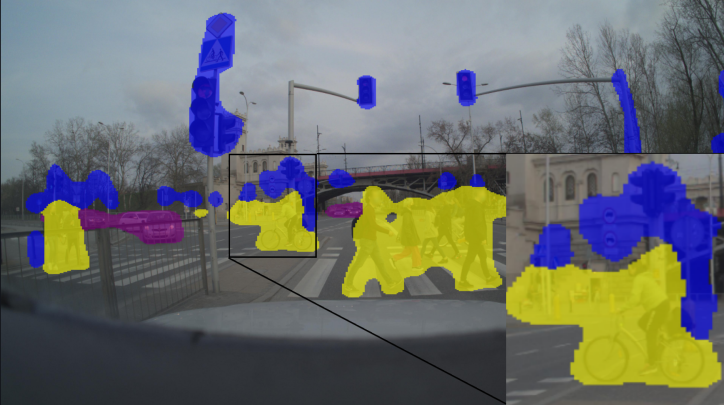}
\subcaption{Baseline CLFT-Hybrid}
\label{fig:qualitative_baseline}
\end{minipage}
\hfill
\begin{minipage}[b]{0.32\textwidth}
\centering
\includegraphics[width=\textwidth]{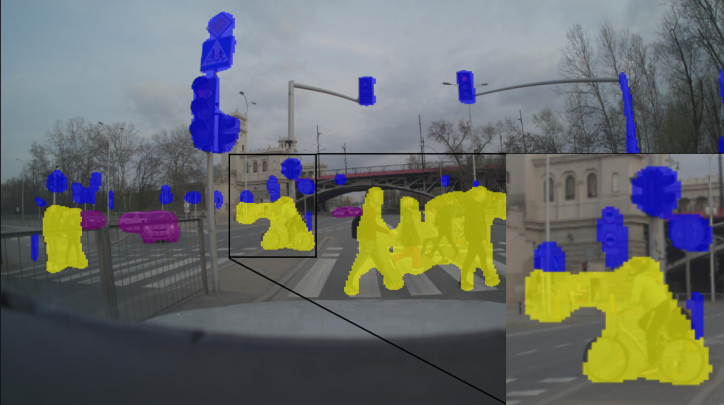}
\subcaption{Ensemble prediction}
\label{fig:qualitative_ensemble}
\end{minipage}

\caption{Qualitative comparison of segmentation results on ZOD frame 000404. (a) Ground truth segmentation with zoomed inset. (b) Baseline prediction showing less precise object boundaries in the zoom. (c) Ensemble prediction demonstrating improved boundary precision in the zoomed area for small safety-critical objects (pedestrians, cyclists, and road signs).}
\label{fig:qualitative_results}
\end{figure*}

\subsection{Annotation Comparison}
\begin{table}[htbp]
\centering
\caption{Validation Set Comparison of Dense (SAM-generated) vs. Sparse (LiDAR-projected) Annotations on ZOD using CLFT-Hybrid (IoU \%)}
\begin{tabular}{@{}llccccc@{}}
\toprule
Modality & Annotation & mIoU & Vehicle & Sign & Human \\
\midrule
\multirow{2}{*}{Camera} & Camera & 39.9 & 63.5 & 24.5 & 31.8 \\
& LiDAR & 24.2 & 36.8 & 15.3 & 20.6 \\
\cmidrule{2-6}
\multirow{2}{*}{LiDAR} & Camera & 29.8 & 49.9 & 18.5 & 21.1 \\
& LiDAR & 21.1 & 32.9 & 14.6 & 15.8 \\
\cmidrule{2-6}
\multirow{2}{*}{Fusion} & Camera & 42.9 & 65.4 & 29.7 & 33.6 \\
& LiDAR & 27.4 & 39.7 & 20.3 & 22.2 \\
\bottomrule
\end{tabular}
\label{tab:annotation_comparison}
\end{table}

Table \ref{tab:annotation_comparison} shows that evaluations using sparse LiDAR annotations yield significantly lower IoU scores.
This drop is expected: the model outputs dense object masks, but the LiDAR ground truth contains only a handful of points per object.
When these dense predictions are compared to sparse labels, most of the predicted object area is treated as background, artificially inflating false positives and reducing IoU.

This discrepancy reflects a supervision mismatch.
Sparse LiDAR annotations provide only limited surface samples, which can lead models to align their predictions with the sampling pattern rather than the full object geometry.
In contrast, the dense SAM‑generated masks offer more complete spatial coverage, enabling camera‑only and fusion models to infer object extent between scan lines using image cues. This provides a supervision signal that more closely approximates the physical object extent than the sparse LiDAR sampling.

\subsection{Model Specialization Analysis}
\begin{table}[htbp]
\centering
\caption{Model Specialization and Ensemble Results on ZOD Validation Set}
\begin{tabular}{@{}l l c c c c @{}}
\toprule
Architecture & Configuration & Vehicle & Sign & Human \\
\midrule
CLFT-Hybrid & Baseline & 65.4 & 29.7 & 33.6 \\
Fusion & Vehicle Only & 70.4 & - & - \\
& Sign Only & - & 43.5 & - \\
& Human Only & - & - & 38.0 \\
& Vehicle+Sign & \textbf{71.0} & 41.7 & - \\
& Vehicle+Human & 69.5 & - & 37.3 \\
& Sign+Human & - & 41.6 & \textbf{40.2} \\
& Ensemble & 70.8 & \textbf{44.0} & 39.2 \\
\midrule
DeepLabV3+ & Baseline & 56.4 & 19.5 & 23.5 \\
Fusion & Vehicle Only & 60.2 & - & - \\
& Sign Only & - & \textbf{26.8} & - \\
& Human Only & - & - & \textbf{30.2} \\
& Vehicle+Sign & 62.3 & 23.9 & - \\
& Vehicle+Human & \textbf{62.8} & - & 29.2 \\
& Sign+Human & - & 24.9 & 28.1 \\
\bottomrule
\end{tabular}
\label{tab:specialization_results_zod}
\end{table}

To address the severe class imbalance in ZOD, we trained specialized models for single classes (Vehicle, Human, Sign) and for pairwise combinations (Table \ref{tab:specialization_results_zod}).
Specialization reduces the optimization scope to a subset loss $L_S(\theta_S) = \sum_{c \in S} w_c L_c(\theta_S)$, allowing the model to allocate capacity more effectively to the selected classes.

Single‑class specialization yields substantial improvements.
For example, the Sign‑only model increases IoU from 29.7\% to 43.5\%, demonstrating that rare classes benefit from dedicated capacity.
This suggests that the shared decoder in the baseline model struggles to represent features for classes with limited training signal.
Small, reflective objects such as signs differ in scale, appearance, and spatial frequency from large, textured objects like vehicles, and a unified feature space may underrepresent these minority classes.

Pairwise specialization shows similar trends: combining related classes (e.g., Sign+Human) improves performance on both, while ensembles further boost results by aggregating complementary strengths across specialized models.

\subsection{Ensemble Method and Efficiency}
To combine the strengths of the specialized models, we construct a parameter‑merged ensemble (Figure \ref{fig:qualitative_results}).
On the ZOD validation set, this ensemble reaches 51.3\% mIoU, outperforming the CLFT‑Hybrid multi‑class baseline.
As shown in Table \ref{tab:computational_efficiency}, this improvement comes with a substantial computational cost: 10.3 FPS compared to the baseline’s 27.8 FPS.

Although the ensemble achieves the highest accuracy among our configurations, its inference speed limits its suitability for real‑time driving.
Instead, it is better positioned as an offline teacher model for auto‑labeling additional data or for distilling its knowledge into more efficient student networks.

\begin{table}[htbp]
\centering
\caption{Fusion Computational Efficiency Comparison on NVIDIA GeForce RTX 5070 Ti 16GB GPU and Intel Core i5-13600K CPU}
\begin{tabular}{@{}l c c c c @{}}
\toprule
Metric & DeepLabV3+ & CLFT-Hybrid & CLFT-Large & Ensemble \\
\midrule
Mean FPS & 53.46 & 27.82 & 15.78 & 10.3 \\
Param. (M) & 118.7 & 127.6 & 341.2 & 382.8 \\
FLOPs (G) & 100.5 & 185.5 & 440.5 & 556.52 \\
GPU (ms) & 18.7 & 35.9 & 63.4 & 93.74 \\
CPU (ms) & 544.7 & 816.2 & 2054.3 & 2568.47 \\
\bottomrule
\end{tabular}
\label{tab:computational_efficiency}
\end{table}

\subsection{Validation on Iseauto Platform}

\begin{figure}[ht]
\centering
\includegraphics[width=0.9\columnwidth]{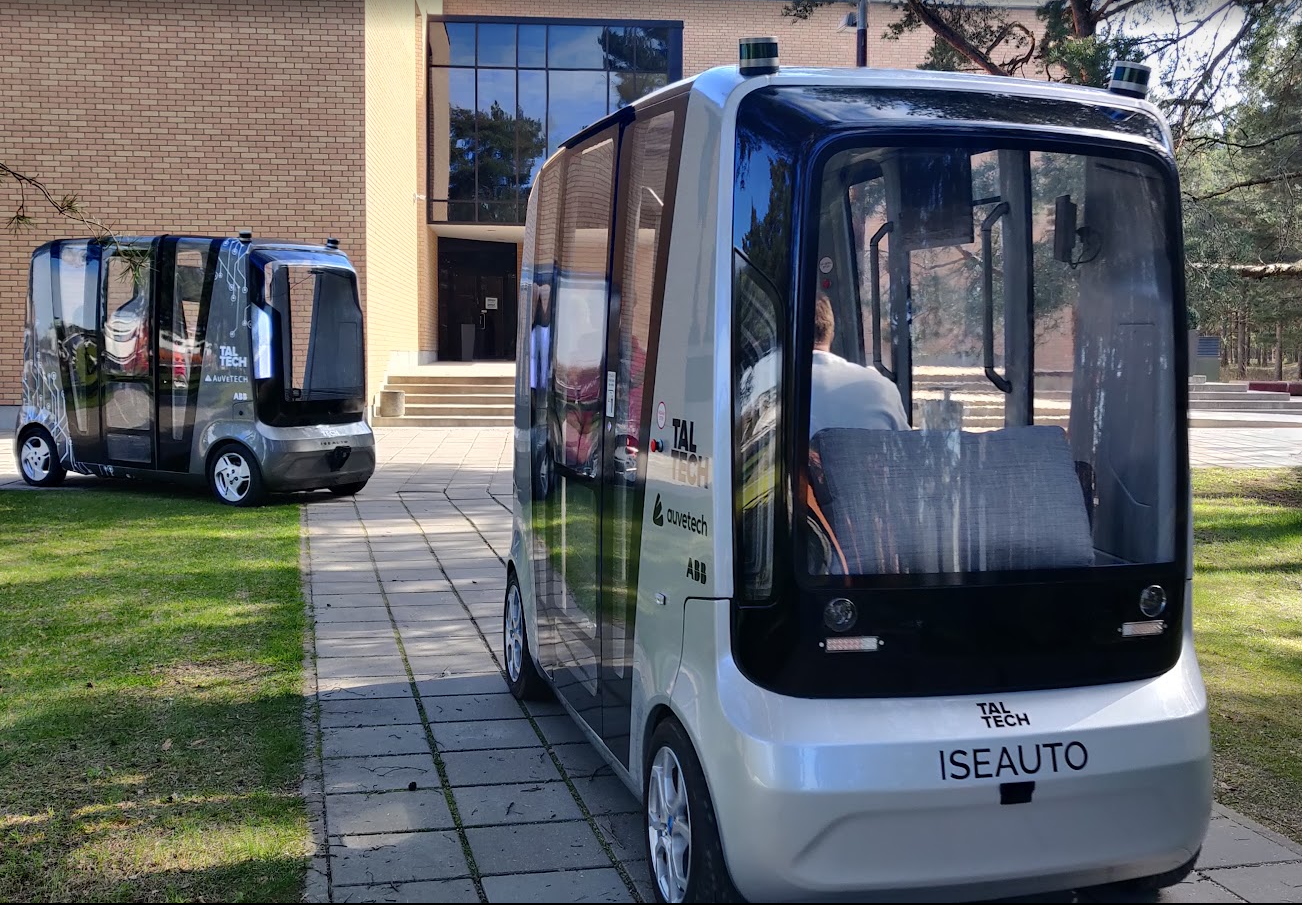}
\caption{Iseauto platform equipped with Velodyne lidars, 4K cameras, IMU, GNSS receiver for data collection and practical application in autonomous driving scenarios.}
\label{fig:iseauto_vehicle}
\end{figure}

To assess the generality of our approach in a real‑world setting, we evaluate on the Iseauto platform (Figure \ref{fig:iseauto_vehicle}).
The results in Table \ref{tab:iseauto_baseline} demonstrate that our method transfers effectively: the CLFT‑Hybrid fusion model reaches 77.5\% mIoU under Day Fair conditions, substantially higher than the 48.1\% mIoU observed on ZOD.

This difference reflects the distinct characteristics of the two datasets.
Iseauto contains clearer object boundaries, fewer overlapping instances, and more structured urban scenes, which reduce ambiguity for both camera and LiDAR modalities.
In contrast, ZOD includes dense traffic, numerous small and distant objects, and more varied environmental conditions, making segmentation inherently more challenging.

Finally, SAM‑enhanced segmentations (Table \ref{tab:edge_fine_tuning}) further improve human IoU in adverse conditions, indicating that denser supervision helps models better capture small, low‑visibility objects in real‑world scenarios.

\begin{table}[htbp]
\centering
\caption{Modality Ablation Study on Iseauto Dataset Baseline Performance Across CLFT Architectures, and Weather Conditions (IoU \%)}
\begin{tabular}{@{}l l c c c c@{}}
\toprule
Architecture & Weather & mIoU & FW IoU & Vehicle & Human \\
\midrule
CLFT-Base & Day Fair & 70.5 & 72.7 & 75.2 & 65.7 \\
Camera-only & Day Rain & 67.4 & 77.9 & 79.2 & 55.5 \\
 & Night Fair & 61.1 & 71.7 & 74.1 & 48.1 \\
 & Night Rain & 51.6 & 58.9 & 60.0 & 43.2 \\
\midrule
CLFT-Base & Day Fair & 61.0 & 63.6 & 66.3 & 55.6 \\
LiDAR-only & Day Rain & 59.3 & 69.9 & 71.2 & 47.4 \\
 & Night Fair & 55.9 & 65.9 & 68.1 & 43.7 \\
 & Night Rain & 49.7 & 58.7 & 60.0 & 39.3 \\
\midrule
CLFT-Base & Day Fair & 72.7 & 75.9 & 79.3 & 66.0 \\
Fusion & Day Rain & 66.9 & 79.0 & 80.5 & 53.2 \\
 & Night Fair & 63.3 & 74.2 & 76.6 & 50.0 \\
 & Night Rain & 60.0 & 65.2 & 66.0 & 53.9 \\
\midrule
\midrule
CLFT-Hybrid & Day Fair & 75.2 & 76.9 & 78.8 & 71.6 \\
Camera-only & Day Rain & 66.3 & \textbf{79.0} & 80.6 & 52.0 \\
 & Night Fair & 66.0 & 73.8 & 75.6 & 56.4 \\
 & Night Rain & 56.5 & 60.2 & 60.8 & 52.1 \\
\midrule
CLFT-Hybrid & Day Fair & 64.0 & 65.5 & 67.1 & 60.9 \\
LiDAR-only & Day Rain & 60.5 & 71.9 & 73.2 & 47.8 \\
 & Night Fair & 57.8 & 68.5 & 70.8 & 44.7 \\
 & Night Rain & 54.6 & 61.2 & 62.2 & 47.0 \\
\midrule
CLFT-Hybrid & Day Fair & \textbf{77.5} & 78.4 & 79.5 & \textbf{75.4} \\
Fusion & Day Rain & \textbf{68.5} & \textbf{79.3} & 80.6 & 56.4 \\
 & Night Fair & 65.8 & 75.4 & 77.5 & 54.0 \\
 & Night Rain & 60.3 & 66.2 & 67.1 & 53.5 \\
\midrule
\midrule
CLFT-Large & Day Fair & 75.6 & 76.7 & 78.0 & 73.1 \\
Camera-only & Day Rain & 65.3 & 77.3 & 78.8 & 51.8 \\
 & Night Fair & 66.1 & 73.6 & 75.3 & 56.9 \\
 & Night Rain & 55.2 & 61.3 & 62.2 & 48.1 \\
\midrule
CLFT-Large & Day Fair & 62.0 & 63.9 & 66.0 & 58.0 \\
LiDAR-only & Day Rain & 60.1 & 67.4 & 68.3 & 51.8 \\
 & Night Fair & 56.4 & 66.3 & 68.6 & 44.1 \\
 & Night Rain & 51.5 & 58.2 & 59.1 & 43.8 \\
\midrule
CLFT-Large & Day Fair & 76.3 & 78.4 & \textbf{80.7} & 71.8 \\
Fusion & Day Rain & \textbf{68.5} & \textbf{79.3} & 80.6 & 56.4 \\
 & Night Fair & \textbf{66.4} & 75.9 & 78.1 & 54.6 \\
 & Night Rain & \textbf{60.6} & 64.4 & 65.0 & 56.2 \\
\bottomrule
\end{tabular}
\label{tab:iseauto_baseline}
\end{table}

\begin{table}[t]
\centering
\caption{Impact of SAM-Enhanced Edge Fine-Tuning on Segmentation Performance in Adverse Conditions on Iseauto Dataset (Camera modality IoU results)}
\label{tab:edge_fine_tuning}
\begin{tabular}{@{}l c c c c@{}}
\toprule
Condition & Metric & Original & Enhanced & Improvement \\
\midrule
\multirow{2}{*}{Night Fair} & Human IoU & 53.1 & 54.5 & +1.4 \\
& Vehicle IoU & 74.9 & 74.2 & -0.7 \\
\midrule
\multirow{2}{*}{Day Rain} & Human IoU & 52.3 & 55.2 & +2.9 \\
& Vehicle IoU & 78.4 & 79.0 & +0.6 \\
\midrule
\multirow{2}{*}{Night Rain} & Human IoU & 41.4 & 46.7 & +5.3 \\
& Vehicle IoU & 60.7 & 59.4 & -1.3 \\
\midrule
\multirow{2}{*}{Average} & Human IoU & 48.9 & 52.1 & +3.2 \\
& Vehicle IoU & 71.3 & 70.9 & -0.4 \\
\bottomrule
\end{tabular}
\end{table}

\subsection{Bidirectional Transfer Learning Analysis}
\begin{figure*}[h]
    \centering
    \begin{tikzpicture}
        \begin{axis}[
            title={Validation Mean IoU Over Epochs (Bidirectional Transfer Learning)},
            xlabel={Epoch},
            ylabel={Mean IoU},
            legend pos=south east,
            legend style={font=\footnotesize},
            grid=major,
            width=\textwidth,
            height=0.4\textwidth,
            xmin=0,
            xmax=200,
            ymin=0.1,
            ymax=0.75
        ]
        \addplot[color={rgb:red,0;green,0;blue,0.6}, line width=1.5pt, mark=none] coordinates {
            (0,0.1763)
            (20,0.1862)
            (40,0.2359)
            (60,0.3070)
            (80,0.3494)
            (100,0.3623)
            (120,0.3713)
            (140,0.3979)
            (160,0.4240)
            (180,0.4246)
            (200,0.4423)
        };
        \addlegendentry{ZOD Baseline}
        \addplot[color={rgb:red,0;green,0.2;blue,1}, dashed, line width=1.5pt, mark=none] coordinates {
            (0,0.2034)
            (20,0.2360)
            (40,0.2496)
            (60,0.2944)
            (80,0.3552)
            (100,0.3629)
            (120,0.3948)
            (140,0.4150)
            (160,0.4176)
            (180,0.4255)
            (200,0.4419)
        };
        \addlegendentry{ISEAuto $\rightarrow$ ZOD LR 8e-5}
        \addplot[color={rgb:red,0.4;green,0.7;blue,1}, dotted, line width=1.5pt, mark=none] coordinates {
            (0,0.1572)
            (20,0.2299)
            (40,0.2946)
            (60,0.3211)
            (80,0.3751)
            (100,0.4072)
            (120,0.4089)
            (140,0.4299)
            (160,0.4421)
            (180,0.4604)
            (200,0.4670)
        };
        \addlegendentry{ISEAuto $\rightarrow$ ZOD LR 5e-5}
        \addplot[color={rgb:red,0;green,0.5;blue,0}, line width=1.5pt, mark=none] coordinates {
            (0,0.2629)
            (20,0.4566)
            (40,0.5529)
            (60,0.6322)
            (80,0.6269)
            (100,0.6683)
            (120,0.6912)
            (140,0.6879)
            (160,0.7099)
            (180,0.7152)
            (200,0.7273)
        };
        \addlegendentry{ISEAuto Baseline}
        \addplot[color={rgb:red,0;green,1;blue,0}, dashdotted, line width=1.5pt, mark=none] coordinates {
            (0,0.1725)
            (20,0.5755)
            (40,0.6039)
            (60,0.6547)
            (80,0.6682)
            (100,0.6786)
            (120,0.7071)
            (140,0.7066)
            (160,0.7152)
            (180,0.7133)
            (200,0.7148)
        };
        \addlegendentry{ZOD $\rightarrow$ ISEAuto LR 8e-6}
        \addplot[color={rgb:red,0.5;green,1;blue,0.5}, densely dashed, line width=1.5pt, mark=none] coordinates {
            (0,0.3105)
            (20,0.6153)
            (40,0.6305)
            (60,0.6630)
            (80,0.6744)
            (100,0.6815)
            (120,0.7085)
            (140,0.7086)
            (160,0.7214)
            (180,0.7213)
            (200,0.7157)
        };
        \addlegendentry{ZOD $\rightarrow$ ISEAuto LR 1e-5}
        \end{axis}
    \end{tikzpicture}
    \caption{Validation Mean IoU trajectories demonstrating accelerated convergence through bidirectional transfer learning between ZOD and ISEAuto datasets. At epoch 20, ISEAuto→ZOD increases mIoU from 18.6\% to 23.6\%, and ZOD→ISEAuto increases mIoU from 45.7\% to 57.6\% . Trained for 200 epochs using CLFT‑Base (RGB) with different learning rates.}
    \label{fig:bidirectional_transfer_iou}
\end{figure*}
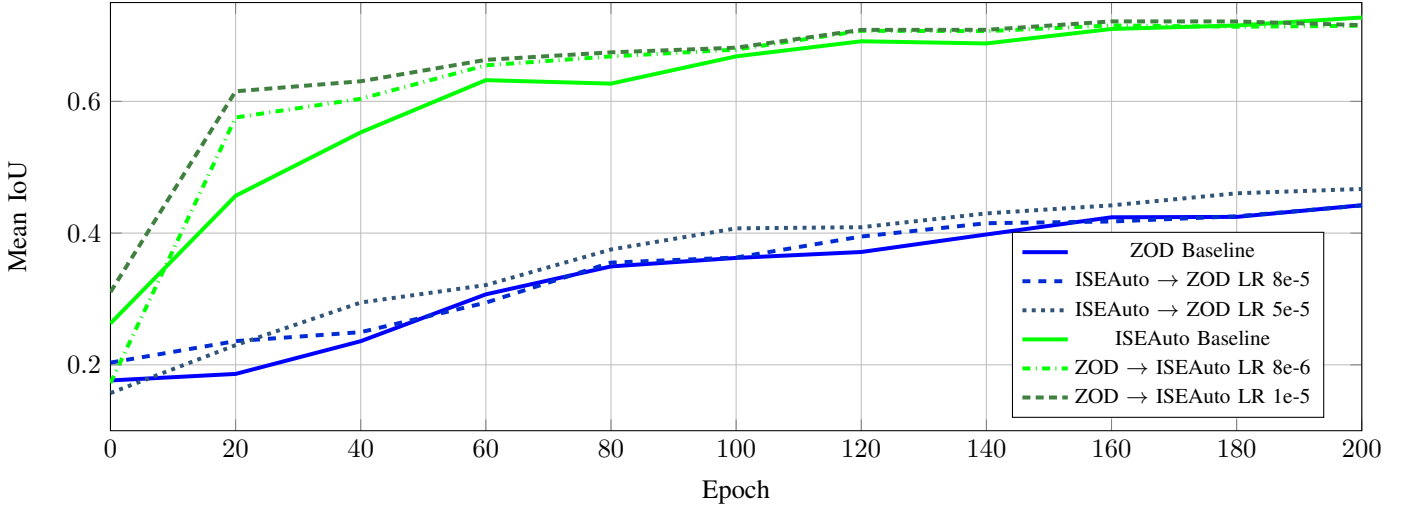

To assess cross‑domain transferability between ZOD and Iseauto, we conduct bidirectional transfer learning experiments using reduced learning rates ($10^{-6}$ to $3 \times 10^{-5}$).
As shown in Figure \ref{fig:bidirectional_transfer_iou}, transfer learning accelerates convergence in both directions: models initialized from the opposite dataset reach competitive performance within the first 20–30 epochs, substantially faster than training from scratch.

For ZOD to Iseauto transfer, the transferred models start with higher initial mIoU and converge to final performance within 1–2\% of the baseline trained directly on Iseauto.
Similarly, Iseauto to ZOD transfer improves early‑stage learning dynamics, narrowing the performance gap relative to scratch training over the first 100 epochs.

These results indicate that the SAM‑enhanced datasets provide feature representations that generalize across domains with differing scene structure and object distributions.
While the final performance remains dataset‑dependent, the accelerated convergence suggests that the learned representations capture transferable geometric and semantic cues useful for autonomous‑driving perception.

\section{Availability}
All components of our pipeline are released as open source to support reproducible research. The SAM‑based annotation generator is available at
\url{https://github.com/taltech-av/paper-aim2026-zod-sam-generator},
and the multi‑modal fusion training framework is provided at
\url{https://github.com/taltech-av/paper-aim2026-fusion-trainer}.
Processed datasets used in this work, including the curated SAM‑generated ZOD segmentation subset and the SAM‑enhanced Iseauto annotations, can be downloaded from
\url{https://app.visin.eu/datasets}.
Training logs, model outputs, and visualization dashboards are accessible at
\url{https://app.visin.eu/projects/sam-zod}.

\section{Conclusion}

This work introduces dense multi‑modal object segmentation for the ZOD dataset, enabling pixel‑level perception of dynamic agents on a benchmark previously constrained to sparse LiDAR annotations or road-only labels.
Through a SAM‑based annotation pipeline and a systematic evaluation of transformer‑based CLFT models and CNN‑based DeepLabV3+, we provide the first comprehensive analysis of dense segmentation performance on ZOD, achieving up to 48.1\% mIoU in fusion settings.

We further validated the approach on the Iseauto autonomous vehicle dataset, where SAM‑refined manual annotations enabled the CLFT‑Hybrid model to reach 77.5\% mIoU in real‑world conditions.
Additional experiments explored model specialization and parameter‑merged ensembles, highlighting accuracy gains alongside computational trade‑offs.
Finally, bidirectional transfer learning between ZOD and Iseauto demonstrated accelerated convergence and strong cross‑domain initialization, indicating that dense SAM‑enhanced labels support transferable feature learning across diverse driving environments.

Despite these advances, several challenges remain.
SAM‑based masks can introduce inaccuracies for small, distant, or occluded objects, and transformer architectures incur higher computational costs than CNNs.
Performance degradation in adverse weather and nighttime conditions also raises the need for improvement and domain adaptation.

Future work should focus on scaling the curated ZOD subset, improving small‑object and safety‑critical class segmentation, and broadening validation to additional autonomous‑driving datasets.
It should also examine newer foundation models, pseudo‑labeling strategies, and more efficient ensemble methods for deployment‑oriented settings.

\section*{Acknowledgement}
This research has received funding from the European Union’s Horizon 2020 Research and Innovation Programme under grant agreement No. 856602 (Finest Twins), from the European Union's Horizon Europe Research and Innovation Programme under grant agreement No.~101135988 (PLIADES: AI-Enabled Data Lifecycles Optimization and Data Spaces Integration for Increased Efficiency and Interoperability).

\bibliographystyle{IEEEtran}
\bibliography{references}

\end{document}